\documentclass[conference]{IEEEtran}
\IEEEoverridecommandlockouts
\usepackage{cite}
\usepackage{amsmath,amssymb,amsfonts}
\usepackage{algorithmic}
\usepackage{multirow}
\usepackage{graphicx}
\usepackage{textcomp}
\usepackage{xcolor}
\usepackage{colortbl}
\def\BibTeX{{\rm B\kern-.05em{\sc i\kern-.025em b}\kern-.08em
    T\kern-.1667em\lower.7ex\hbox{E}\kern-.125emX}}
\begin{document}

\title{PheME: A deep ensemble framework for improving phenotype prediction from multi-modal data
\footnotesize
\thanks{* denotes equal contributions}
}


\author{\IEEEauthorblockN{Shenghan Zhang$^1$ *, Haoxuan Li$^2$ *, Ruixiang Tang$^3$ *, Sirui Ding$^4$ *, Laila Rasmy$^5$, Degui Zhi$^5$, Na Zou$^4$, Xia Hu$^3$}
\IEEEauthorblockA{\textit{$^1$ Xi'an Jiaotong-Liverpool University; $^2$ Wuhan University; $^3$ Rice University;
    $^4$ Texas A\&M University; } \\ \textit{$^5$ University of Texas Health Science Center, School of Biomedical Informatics;}
    }
}

\author{\hspace{-30pt}\IEEEauthorblockN{1\textsuperscript{st} Shenghan Zhang*}
\IEEEauthorblockA{
\hspace{-30pt}\textit{Department of Computer Science} \\
\hspace{-30pt}\textit{Xi'an Jiaotong-Liverpool University}\\
\hspace{-30pt}Xi'An, China \\
\hspace{-30pt}Shenghan.Zhang19@student.xjtlu.edu.cn }
\and
\IEEEauthorblockN{2\textsuperscript{nd} Haoxuan Li*}
\IEEEauthorblockA{
\textit{Department of Computer Science} \\
\textit{Wuhan University}\\
Wuhan, China \\
2019302180080@whu.edu.cn}
\and
\IEEEauthorblockN{3\textsuperscript{rd} Ruixiang Tang*}
\IEEEauthorblockA{
\textit{Department of Computer Science} \\
\textit{Rice University}\\
Houston, USA \\
rt39@rice.edu}
\and
\IEEEauthorblockN{4\textsuperscript{th} Sirui Ding*}
\IEEEauthorblockA{
\textit{Department of Computer Science } 
 \\ {and Engineering} \\
\textit{Texas A\&M University}\\
College Station, USA \\
siruiding@tamu.edu} 
\and
\IEEEauthorblockN{5\textsuperscript{th} Laila Rasmy}
\IEEEauthorblockA{\textit{UTHealth Houston School} \\{of Biomedical Informatics  }\\ 
Houston, USA \\
Laila.Rasmy.GindyBekhet@uth.tmc.edu}
\and
\IEEEauthorblockN{6\textsuperscript{th} Degui Zhi}
\IEEEauthorblockA{\textit{UTHealth Houston School} \\{of Biomedical Informatics  }\\
Houston, USA \\
Degui.Zhi@uth.tmc.edu}
\and
\IEEEauthorblockN{7\textsuperscript{th} Na Zou}
\IEEEauthorblockA{\textit{Department of Engineering Technology} \\ {\& Industrial Distribution} \\
\textit{Texas A\&M Univerisity}\\
College Station, USA \\
nzou1@tamu.edu}
\and
\hspace{-150pt}
\IEEEauthorblockN{8\textsuperscript{th} Xia Hu}
\hspace{-130pt}\IEEEauthorblockA{ \hspace{-130pt} \textit{Deparment of Computer Science} \\
\hspace{-130pt} \textit{Rice University}\\
\hspace{-130pt} Houston, USA \\
\hspace{-130pt} xia.hu@rice.edu} \\
}

\maketitle

\begin{abstract}
Detailed phenotype information is fundamental to accurate diagnosis and risk estimation of diseases. As a rich source of phenotype information, electronic health records (EHRs) promise to empower diagnostic variant interpretation. However, how to accurately and efficiently extract phenotypes from the heterogeneous EHR data remains a challenge. In this work, we present PheME, an Ensemble framework using Multi-modality data of structured EHRs and unstructured clinical notes for accurate Phenotype prediction. Firstly, we employ multiple deep neural networks to learn reliable representations from the sparse structured EHR data and redundant clinical notes. A multi-modal model then aligns multi-modal features onto the same latent space to predict phenotypes. Secondly, we leverage ensemble learning to combine outputs from single-modal models and multi-modal models to improve phenotype predictions. We chose seven diseases to evaluate the phenotyping performance of the proposed framework. Experimental results show that using multi-modal data significantly improves phenotype prediction in all diseases, the proposed ensemble learning framework can further boost the performance.
\end{abstract}

\begin{IEEEkeywords}
Phenotype Prediction, Multi-Modal Data, Ensemble Learning, Natural Language Processing
\end{IEEEkeywords}

\section{Introduction}
Throughout the years, clinicians routinely document the care of patients in electronic health records (EHRs) and accumulate a huge amount of medical records~\cite{shickel2017deep, tang2023does, chuang2023spec, yuan2023llm, chang2023towards}. There is a growing interest in utilizing the EHRs to identify detailed phenotypes for disease diagnosis and research purposes~\cite{morley2014defining, tang2019classification, rizvi2023local}. However, extracting phenotype information from EHRs is not an easy task~\cite{pathak2013electronic} due to the data fragmentation, multi-modality, and lack of uniform inclusion criteria. The phenotype evidence can derive from the structured International Classification of Diseases, 9th and 10th revision (ICD-9/10) codes~\cite{slee1978international,who1992international}, longitudinal lab testing results, or unstructured clinical notes documented by physicians. Due to the varying characteristics of the disease, different data modalities can be better or worse at intensifying specific phenotypes. For example, a majority of breast cancer patients are identifiable with their ICD codes or medications. In terms of classifying specific stages of breast cancer which is determined by the size and shape of the tumor, clinical notes can be a more useful resource~\cite{zhou2021cancerbert}. Besides, EHRs are known to include coding and reporting biases from doctors~\cite{agniel2018biases}, presenting further challenges to identifying patient cohorts \cite{tang2021mitigating}. Therefore, identifying patients' phenotypes from the noisy and multi-modal EHRs remains a challenging problem.

EHR-based phenotyping is a classification task to identify certain medical diagnoses with the patient's EHRs. In previous works, researchers try to solve this problem by building plenty of rule-based algorithms~\cite{kirby2016phekb}, such as specifying the abnormal value of laboratory testing results, inclusion or exclusion of certain ICD codes, or the existence of disease-related phrases in clinical notes. Rule-based phenotyping algorithms, e.g., Phenotype KnowledgeBase (PheKB)~\cite{kirby2016phekb} are manually built by researchers with advanced medical knowledge of the disease. However, due to the limitation of expert knowledge, the rule-based performance may vary widely when applied to different hospital EHR systems and patients~\cite{de2021phe2vec}. In addition, the missing data, input errors, and reporting bias further degrade the accuracy of rule-based algorithms in real-world applications. For example, rules often fail when certain phrases or keywords are not explicitly stated in the clinical notes.

Automated phenotyping with machine learning provides an alternative that could be more generalized and scalable compared to rule-based algorithms~\cite{de2021phe2vec}. Afshar et al.~\cite{afshar2020taste} apply the tensor factorization to phenotyping tasks on structured EHRs. Suesh et al.~\cite{suresh2017use} use autoencoders to create low-dimensional embeddings of underlying patient phenotypes and study how different patients will react to different interventions. Allen et al.~\cite{allen2021interpretable} propose the notion of interpretable phenotype and introduce an unsupervised learning framework that can facilitate general clinical validation, and alleviate the problem of high-dimensionality. However, these works focus on the single-modal data, e.g., structured EHRs, without taking full advantage of the multi-modal EHR data including clinical notes. There are some works that apply multi-modality learning to some specific diseases. Kim et al~\cite{kim2020multimodal} propose a multi-modal method based on matrix factorization with MRI and cognitive scores data for the Alzheimer's disease phenotype. Slaby et al.~\cite{slaby2022electronic} use a rule-based algorithm combined with a text mining technique for ADHD phenotyping. Some initial efforts~\cite{wang2022graph,song2022automatic,ahuja2022mixehr} are put into developing the general multi-modal phenotyping framework for various disease types. However, they are based on the topic model which may get unsatisfactory performance on large-scale data~\cite{zhao2021topic}.  


There are several challenges when designing a general deep learning phenotyping framework. We summarize the challenges from two perspectives. From the data perspective, the multi-modality medical data is usually noisy and sparse to learn from. For example, the clinical notes are very long containing redundant unrelated information and the structured EHRs can be very sparse. The fixed context length token of state-of-the-art NLP models like the transformer model~\cite{vaswani2017attention} makes them infeasible to be directly applied to clinical notes. Directly training the model on sparse EHRs will lead to ineffective learning. From the model perspective, we observe from the preliminary experiments that there is no single model that can always achieve satisfying performances on all diseases. This is because diseases may rely on dissimilar diagnostic features and different models have distinct preferences for the input data. Hence, making various models complement each other is a problem that needs to be solved.


To tackle the above challenges, we present an ensemble learning framework based on multi-modal data for EHR-based phenotyping. The inputs of the model consist of structured EHRs and clinical notes. For structured EHRs, we adopt a multi-layer perception to project the sparse tabular data into a dense embedding. For the clinical notes data, we utilize a text filter to select important sentences that contain disease-related phrases or keywords and feed them into the language model pre-trained on the medical domain. In this way, we can obtain uniform representations for the two modalities. Then, we use a deep learning module to encourage information sharing between two modalities and integrate the representations from both modalities for phenotype prediction. In addition, different from previous studies that only train a single model for prediction, we train multiple models and use multiple model fusion techniques to reduce the prediction error. Our results show that by effectively learning from the multi-modality data with the proposed multi-modal framework, we improve the phenotype prediction performance over seven diseases by $14.28\%$ on ROC AUC compared to the multi-modality baseline model. In addition, our ensemble learning framework further boosts the multi-modality model's performance by $4.28\%$ improvement in ROC AUC with the average of two ensemble strategies. For diseases with limited patient samples such as sickle cell, the ensemble framework receives a significant improvement of ROC AUC from $0.771$ to $0.968$ compared to the multi-modality baseline model. This provides a further potential application for some rare disease phenotyping tasks.

\section{Method}
\begin{figure*}[]
\centering
\includegraphics[width=0.8\textwidth]{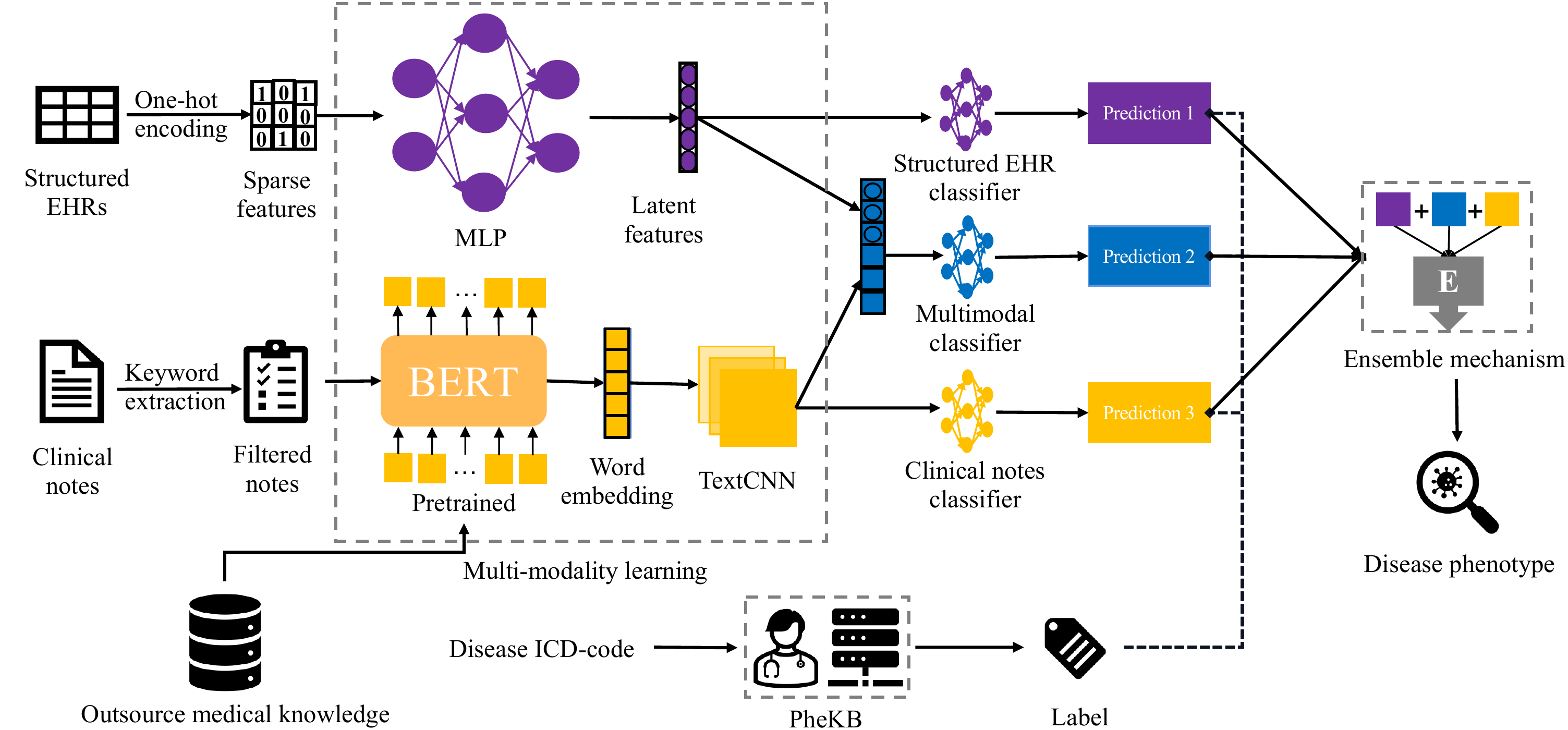}
 \caption{An overview of the workflow for our deep ensemble framework}
 \label{fig:workflow}
\end{figure*}

In this section, we discuss the details of the proposed framework. Firstly, we will introduce the problem formulation and notations used in this work. We then elaborate on the multi-modality learning from both structured EHR data and clinical notes data. Finally, we present the ensemble learning framework with two ensemble strategies. The overview of the workflow is presented in Figure~\ref{fig:workflow} as follows.


\subsection{Problem formulation}

Given a structured EHR data $\mathbf{x^s}$ and clinical notes $\mathbf{x^t}$ from a patient, our goal is to build a classification model $ \mathbf{f(x^s, x^t)}$ and predict phenotype $\hat{y}$ from the targeted diseases (we select seven diseases in this study), where $\hat{y} \in \{0, 1\}$ and $1$ specifies the existence of the disease. Since most of the EHR datasets lacking explicit phenotype label information, we adopt the PheKB, a rule-based phenotyping system, to provide approximate training labels $y=\mathbf{PheKB(x^s, x^t)}$ and train multiple classification models $\mathbf{f_i(x^s,x^t)}, i \in [1,...,n]$ with the weakly-supervised data  $D = \mathbf{\{(x^s,x^t),} y\}$. To make the system more generalized and scalable, we leverage ensemble learning and integrate the outputs from multiple models, i.e., $ \mathbf{f(x^s, x^t) = E(f_i(x^s, x^t))}, i \in [1,..,n]$, where $\mathbf{E}$ specifies the ensemble approach. To evaluate the model's performance, we first evaluate the model by calculating the metrics e.g., ROC AUC and F1 Score, based on the predicted output $\hat{y}$ and pseudo label $y$. To further validate our model's generalization ability, we also select some patients and include medical experts to manually check the model's performance.

\subsection{Weak supervised setting}
Electronic health record (EHR) data has long been used for disease phenotyping. However, it is infeasible for researchers to reliably identify patients with certain diseases from EHRs due to the lack of specific inclusion criterias. Only researchers with medical knowledge and advanced research on certain diseases are capable to select patient cohorts. All of these challenges make it a problem for us to filter out patients with specific diseases from the EHR data to conduct experiments. Fortunately, Phenotype KnowledgeBase(PheKB) \cite{kirby2016phekb}, established by the Electronic Medical Records and Genomics (eMERGE), creates and defines many rule-based phenotype algorithms based on inclusion of certain ICD codes or phrases for identification. While these algorithms could be reliable with enough complete and valid information of patients, there is a limited scalability of them due to the excessively detailed design base on the knowledge of experts. In addition, the missing data, input errors, and reporting bias made by physicians further degrade the performance of rule-based algorithms. Hence, we use the patient cohorts selected by PheKB as a pseudo-label for the weak supervised learning.



\subsection{Multi-modality learning}
To maximize the use of patients' data, we propose a multi-modality learning framework to effectively extract phenotype information from EHRs. Specifically, a multi-layer perceptron and language model are adopted to learn data representations from the structured EHRs and clinic notes, respectively. We introduce the details as follows. 

\textbf{Learning from structured EHR data.}
Structured EHR data consist of medical codes, including medications, laboratory data, etc.  To efficiently extract phenotype related features from the large and sparse dataset, we use one-hot encoding to encode different types of ICD codes, medical codes and demographics. For laboratory test results, we only consider the abnormal value. We then employ a multilayer perceptron (MLP) to learn dense representation as follows.
\begin{equation}
    \mathbf{V}^s = \text{MLP}(\mathbf{x}^s),
\end{equation}
where $\mathbf{x}^s$ denotes the one-hot embedded patients' EHR data, $\text{MLP}$ indicates the multilayer perceptron network and $\mathbf{V}^s$ denotes the latent features learnt by MLP from the structured EHR data.  

\textbf{Learning from clinical notes.} The clinical notes contain various nuanced information, e.g, history of patient’s illness and medications. However, the notes are known to be very noisy and contain redundant unrelated information. In Table~\ref{datasetstat}, we observe that the average token number of the notes are more than few thousands. Considering that the common input length of deep language models are 512, it would be infeasible for the direct application of language model to extract useful knowledge from the clinical notes. To address the challenge, we employ a simple yet effective method to extract highly related information from clinical notes as follows.
\begin{equation}
    \mathbf{x'}^t = \text{NoteExtr}(\mathbf{x}^t, \mathbf{k}^t),
\end{equation}
where $\mathbf{x}^t$ denotes the clinical notes of the patients, $\mathbf{k}^t$ specifies the keyword related to the target diseases and $\mathbf{x'}^t$ indicates the extracted clinical notes. The $\text{NoteExtr}$ function conducts the extraction by choosing the target sentence containing the keywords as well as the adjacent sentences.

Because of the high complexity of clinical text data, we decide to leverage the outsource medical knowledge to effectively extract clinical notes' features. Recently, the development of pre-trained language models on large-scale text data, e.g., BERT, has significantly improved natural language processing tasks within the general language domain~\cite{devlin2018bert,min2021recent}. However, in the clinical domain, the vocabulary, syntax, and semantics differ significantly from the general language. Here, we employ the BlueBERT~\cite{peng2019transfer} which is pre-trained on the large volume of medical texts including 4 billion words from PubMed abstracts and 500 million words from MIMIC-III. To aggregate the features learned from the long clinical notes, we employ TextCNN~\cite{zhang2015textcnn} to 
further generate latent representations from the long text into one embedding vector. The whole text process pipeline can be formulated as follows:
\begin{gather}
    \mathbf{V'}^t = \text{BERT}(\mathbf{x'}^t),
    \\
    \mathbf{V}^t = \text{TextCNN}(\mathbf{V'}^t)
\end{gather}
where $\mathbf{V'}^t$ is the embedding of $512$-word chunks from clinical notes with shape $N\times 768$ ($N$ denotes the maximum size of patients' 512-word chunks), $\mathbf{V}^t$ is the final representation extracted by $\text{TextCNN}$ with the length of $384$.


\textbf{Fusing multi-modality representations.} To jointly mine from multi-modality data, we concatenate two representations together for the disease phenotyping as follows.
\begin{gather}
    \mathbf{V}^{s+t} = \text{concact}(\mathbf{V}^s, \mathbf{V}^t),
    \\
    \hat{y} = \text{MLP}(\mathbf{V}^{s+t}),
\end{gather}
where $\mathbf{V}^{s+t}$ denotes the concatenated latent representations of structured data and clinical notes. $\hat{y}$ indicates the prediction of the disease phenotype. 


\begin{table*}[t]
\centering
\caption{Dataset statistics}
\begin{tabular}{
>{\columncolor[HTML]{FFFFFF}}c |
>{\columncolor[HTML]{FFFFFF}}c |
>{\columncolor[HTML]{FFFFFF}}c |
>{\columncolor[HTML]{FFFFFF}}c |
>{\columncolor[HTML]{FFFFFF}}c |
>{\columncolor[HTML]{FFFFFF}}c |
>{\columncolor[HTML]{FFFFFF}}c |
>{\columncolor[HTML]{FFFFFF}}c |}
\hline
{\color[HTML]{343434} \textbf{}} &
  {\color[HTML]{343434} \textbf{Dementia}} &
  {\color[HTML]{343434} \textbf{\begin{tabular}[c]{@{}c@{}}Herpes\\ Zoster\end{tabular}}} &
  {\color[HTML]{343434} \textbf{Asthma}} &
  {\color[HTML]{343434} \textbf{\begin{tabular}[c]{@{}c@{}}Post Event\\ Pain\end{tabular}}} &
  {\color[HTML]{343434} \textbf{\begin{tabular}[c]{@{}c@{}}Sickle\\ Cell\end{tabular}}} &
  {\color[HTML]{343434} \textbf{ADHD}} &
  {\color[HTML]{343434} \textbf{\begin{tabular}[c]{@{}c@{}}Heart\\ Failure\end{tabular}}} \\ \hline
{\color[HTML]{343434} \textbf{Patient Cohorts}} &
  {\color[HTML]{343434} 2110} &
  {\color[HTML]{343434} 142} &
  {\color[HTML]{343434} 2428} &
  {\color[HTML]{343434} 185} &
  {\color[HTML]{343434} 169} &
  {\color[HTML]{343434} 158} &
  {\color[HTML]{343434} 10109} \\ \hline
{\color[HTML]{343434} \textbf{Negative Samples}} &
  {\color[HTML]{343434} 2110} &
  {\color[HTML]{343434} 141} &
  {\color[HTML]{343434} 2428} &
  {\color[HTML]{343434} 185} &
  {\color[HTML]{343434} 169} &
  {\color[HTML]{343434} 158} &
  {\color[HTML]{343434} 10109} \\ \hline
\textbf{Dimension of Tabular Data} &
  2703 &
  1540 &
  3069 &
  1505 &
  1504 &
  1383 &
  4231 \\ \hline
\textbf{Average Tokens of Text Data} &
  7487 &
  10296 &
  7605 &
  5734 &
  6460 &
  8378 &
  8092 \\ \hline
\textbf{Maximum Tokens of Text Data} &
  46080 &
  44544 &
  46592 &
  44032 &
  45056 &
  43008 &
  47104 \\ \hline
\end{tabular}
\label{datasetstat}
\end{table*}

\subsection{Ensemble learning}
Previous studies usually adopt one single model for the prediction. However, our preliminary experimental results show that no single baseline model can achieve competitive performances in all diseases. For example, with the structured EHRs as input, the gradient boosting classifier performs well on the disease of dementia, asthma, sickle cell, and heart failure. But it performs less satisfying on herpes zoster, post event pain, and ADHD. This is because diseases may rely on dissimilar diagnostic features and thus models can have unstable performance. We know that, for some multifactorial diseases, several doctors will get involved in the discussion for the diagnosis and treatment. Motivated by this, we employ the ensemble learning strategy~\cite{sagi2018ensemble} to further boost prediction accuracy. Specifically, multiple models trained with various modality data (analogous to doctors with different knowledge backgrounds) will be combined for the final decision with an ensemble strategy. Our goal is to maximize the diversity of the ensemble models since ensemble learning can generate more accurate classification results than a single classifier with the benefits from the performance of the different modalities and the diversity of the errors~\cite{kuncheva2003measures}. This is supported by the observation that varying data modalities can be better or worse at reflecting reliable diagnoses. For example, patients with atrial fibrillation are distinguishable from medication records, whereas, for patients with rheumatoid arthritis, medications are not useful. In this paper, we employ two kinds of ensemble strategies, which will be introduced in detail as follows.

In this work, we select three ensemble classifiers with different model architectures and input data types. As shown in Figure~\ref{fig:workflow}, they are the structured EHR classifier, clinical notes classifier, and multi-modal classifier. These classifiers are based on the MLP model with the latent representations learned from the raw structured EHRs and clinical notes.

We consider two ensemble techniques in this work. The first kind of ensemble strategy is majority voting~\cite{polikar2012ensemble}. We predict the class with the largest sum of votes from classifiers, which can be formulated as follows.
 \begin{gather}
     {N}^{c} = \sum_{i=1}^n \hat{y}_{i}^c, c=1,2,...,m
     \\
     \hat{y} = argmax(\{{N}^{c}\}_{c=1}^m)
 \end{gather}
where $\hat{y}_{i}^c$ denotes the voter result of $i^{th}$ classifier for the $c^{th}$ class, $N^{c}$ represents the vote number of class $c$. The final prediction $\hat{y}$ is the class that corresponds the maximum vote number from $\{N^{c}\}_{c=1}^m$.

One downside of majority voting is that it gives equal weight to each classifier, however, classifiers are not equally reliable. For certain diseases, we found that some classifiers are much more reliable than others. Then our question would be, for each disease, how to find the reliable classifier and assign a higher weight to them. To tackle the challenges, we adopt a label model strategy~\cite{ratner2018training, ratner2016data}. It generates the final prediction by training a label model with the ensemble members' outputs. So the label model can estimate the accuracies and correlations of the classifiers and the process can be represented with the following formulation.
\begin{equation}
    \hat{y} = \text{M}(\{\hat{y}_i=\mathbf{f_i(x^s,x^t)}\}_{i=1}^n),
\end{equation}
where the $\text{M}$ represents the label model. It takes the set of ensemble members prediction as inputs and outputs the final prediction.

\section{Experimental Results and Analysis}
To demonstrate the effectiveness of our method, we aim to answer three research questions as follows.

\begin{itemize}
    \item \textbf{RQ1:} Can the designed framework precisely predict the phenotype of different diseases? (Section \ref{comparewithbaselines})
    \item \textbf{RQ2:} How can the ensemble learning improve the phenotype prediction performance? (Section \ref{ensemblelearning}) 
    \item \textbf{RQ3:} Does the proposed framework have better generalization ability compared to rule-based methods? (Section \ref{erroranalysis})
\end{itemize}

\begin{table*}[]
\caption{Main experiments}
\footnotesize
\centering
\scalebox{1.05}{
\begin{tabular}{|c|c|lll|ll|lc|}
\hline
\multirow{2}{*}{\textbf{Disease}}         & \multirow{2}{*}{\textbf{Metrics}} & \multicolumn{3}{c|}{\textbf{Structured EHR}}                                                   & \multicolumn{2}{c|}{\textbf{Clinical Notes}}                & \multicolumn{2}{c|}{\textbf{Multi-modality}}                 \\ \cline{3-9} 
&                                   & \multicolumn{1}{c|}{LR}          & \multicolumn{1}{c|}{MLP}          & \multicolumn{1}{c|}{GBC} & \multicolumn{1}{c|}{MLP}          & \multicolumn{1}{c|}{GBC} & \multicolumn{1}{c|}{Baseline}  & \multicolumn{1}{c|}{Ours} \\ \hline
\multirow{4}{*}{\textbf{Dementia}}        & AUC                               & \multicolumn{1}{l|}{0.775±0.018} & \multicolumn{1}{l|}{0.781±0.023} & 0.787±0.017              & \multicolumn{1}{l|}{0.702±0.101} & 0.796±0.018              & \multicolumn{1}{l|}{0.870±0.029} & \textbf{0.966±0.008}      \\ \cline{2-9} 
& Precision                         & \multicolumn{1}{l|}{0.796±0.025} & \multicolumn{1}{l|}{0.769±0.032} & 0.789±0.023              & \multicolumn{1}{l|}{0.704±0.175} & 0.771±0.072              & \multicolumn{1}{l|}{0.861±0.046} & \textbf{0.963±0.014}      \\ \cline{2-9} 
& F1                                & \multicolumn{1}{l|}{0.775±0.034} & \multicolumn{1}{l|}{0.790±0.025} & 0.790±0.031              & \multicolumn{1}{l|}{0.709±0.085} & 0.798±0.029              & \multicolumn{1}{l|}{0.876±0.013} & \textbf{0.967±0.010}      \\ \cline{2-9} 
& Recall                            & \multicolumn{1}{l|}{0.757±0.057} & \multicolumn{1}{l|}{0.813±0.032} & 0.792±0.042              & \multicolumn{1}{l|}{0.783±0.128} & 0.836±0.036              & \multicolumn{1}{l|}{0.896±0.031} & \textbf{0.971±0.013}      \\ \hline
\multirow{4}{*}{\textbf{Herpes zoster}}   & AUC                               & \multicolumn{1}{l|}{0.670±0.038} & \multicolumn{1}{l|}{0.721±0.086} & 0.701±0.037              & \multicolumn{1}{l|}{0.676±0.099} & 0.654±0.051              & \multicolumn{1}{l|}{0.747±0.062} & \textbf{0.897±0.037}      \\ \cline{2-9} 
& Precision                         & \multicolumn{1}{l|}{0.707±0.067} & \multicolumn{1}{l|}{0.741±0.094} & 0.752±0.129              & \multicolumn{1}{l|}{0.520±0.323} & 0.648±0.151              & \multicolumn{1}{l|}{0.747±0.139} & \textbf{0.882±0.062}      \\ \cline{2-9} 
& F1                                & \multicolumn{1}{l|}{0.671±0.105} & \multicolumn{1}{l|}{0.716±0.091} & 0.690±0.114              & \multicolumn{1}{l|}{0.547±0.282} & 0.626±0.072              & \multicolumn{1}{l|}{0.727±0.122} & \textbf{0.902±0.047}      \\ \cline{2-9} 
& Recall                            & \multicolumn{1}{l|}{0.657±0.156} & \multicolumn{1}{l|}{0.706±0.123} & 0.657±0.151              & \multicolumn{1}{l|}{0.659±0.351} & 0.636±0.095              & \multicolumn{1}{l|}{0.717±0.133} & \textbf{0.924±0.044}      \\ \hline
\multirow{4}{*}{\textbf{Asthma}}          & AUC                               & \multicolumn{1}{l|}{0.688±0.022} & \multicolumn{1}{l|}{0.693±0.026} & 0.705±0.027              & \multicolumn{1}{l|}{0.675±0.091} & 0.786±0.012              & \multicolumn{1}{l|}{0.839±0.013} & \textbf{0.966±0.012}      \\ \cline{2-9} 
& Precision                         & \multicolumn{1}{l|}{0.736±0.041} & \multicolumn{1}{l|}{0.706±0.027} & 0.745±0.030              & \multicolumn{1}{l|}{0.548±0.289} & 0.762±0.069              & \multicolumn{1}{l|}{0.835±0.033} & \textbf{0.966±0.012}      \\ \cline{2-9} 
& F1                                & \multicolumn{1}{l|}{0.677±0.065} & \multicolumn{1}{l|}{0.692±0.058} & 0.694±0.068              & \multicolumn{1}{l|}{0.565±0.288} & 0.788±0.039              & \multicolumn{1}{l|}{0.842±0.038} & \textbf{0.967±0.012}      \\ \cline{2-9} 
& Recall                            & \multicolumn{1}{l|}{0.643±0.116} & \multicolumn{1}{l|}{0.685±0.100} & 0.663±0.115              & \multicolumn{1}{l|}{0.604±0.319} & 0.820±0.012              & \multicolumn{1}{l|}{0.852±0.061} & \textbf{0.969±0.016}      \\ \hline
\multirow{4}{*}{\textbf{Post event pain}} & AUC                               & \multicolumn{1}{l|}{0.705±0.042} & \multicolumn{1}{l|}{0.720±0.032} & 0.662±0.052              & \multicolumn{1}{l|}{0.599±0.060} & 0.745±0.035              & \multicolumn{1}{l|}{0.778±0.026} & \textbf{0.922±0.030}      \\ \cline{2-9} 
& Precision                         & \multicolumn{1}{l|}{0.746±0.069} & \multicolumn{1}{l|}{0.758±0.126} & 0.679±0.123              & \multicolumn{1}{l|}{0.596±0.106} & 0.750±0.148              & \multicolumn{1}{l|}{0.780±0.128} & \textbf{0.929±0.038}      \\ \cline{2-9} 
& F1                                & \multicolumn{1}{l|}{0.709±0.039} & \multicolumn{1}{l|}{0.706±0.069} & 0.654±0.101              & \multicolumn{1}{l|}{0.609±0.077} & 0.722±0.078              & \multicolumn{1}{l|}{0.768±0.072} & \textbf{0.928±0.020}      \\ \cline{2-9} 
& Recall                            & \multicolumn{1}{l|}{0.689±0.090} & \multicolumn{1}{l|}{0.679±0.099} & 0.637±0.108              & \multicolumn{1}{l|}{0.707±0.235} & 0.719±0.058              & \multicolumn{1}{l|}{0.771±0.066} & \textbf{0.928±0.024}      \\ \hline
\multirow{4}{*}{\textbf{Sickle cell}}     & AUC                               & \multicolumn{1}{l|}{0.589±0.047} & \multicolumn{1}{l|}{0.623±0.095} & 0.650±0.046              & \multicolumn{1}{l|}{0.552±0.068} & 0.759±0.057              & \multicolumn{1}{l|}{0.771±0.065} & \textbf{0.930±0.023}      \\ \cline{2-9} 
& Precision                         & \multicolumn{1}{l|}{0.606±0.087} & \multicolumn{1}{l|}{0.645±0.135} & 0.673±0.065              & \multicolumn{1}{l|}{0.327±0.268} & 0.764±0.106              & \multicolumn{1}{l|}{0.770±0.092} & \textbf{0.921±0.053}      \\ \cline{2-9} 
& F1                                & \multicolumn{1}{l|}{0.574±0.063} & \multicolumn{1}{l|}{0.616±0.098} & 0.657±0.025              & \multicolumn{1}{l|}{0.387±0.319} & 0.747±0.066              & \multicolumn{1}{l|}{0.774±0.076} & \textbf{0.930±0.020}      \\ \cline{2-9}  
& Recall                            & \multicolumn{1}{l|}{0.554±0.075} & \multicolumn{1}{l|}{0.603±0.103} & 0.655±0.075              & \multicolumn{1}{l|}{0.482±0.410} & 0.752±0.106              & \multicolumn{1}{l|}{0.781±0.079} & \textbf{0.942±0.028}      \\ \hline
\multirow{4}{*}{\textbf{ADHD}}            & AUC                               & \multicolumn{1}{l|}{0.717±0.028} & \multicolumn{1}{l|}{0.764±0.036} & 0.676±0.045              & \multicolumn{1}{l|}{0.637±0.064} & 0.722±0.042              & \multicolumn{1}{l|}{0.769±0.047} & \textbf{0.857±0.074}      \\ \cline{2-9} 
& Precision                         & \multicolumn{1}{l|}{0.731±0.058} & \multicolumn{1}{l|}{0.768±0.104} & 0.679±0.091              & \multicolumn{1}{l|}{0.697±0.193} & 0.706±0.131              & \multicolumn{1}{l|}{0.791±0.124} & \textbf{0.837±0.073}      \\ \cline{2-9} 
                                          & F1                                & \multicolumn{1}{l|}{0.727±0.069} & \multicolumn{1}{l|}{0.756±0.072} & 0.696±0.083              & \multicolumn{1}{l|}{0.546±0.168} & 0.696±0.055              & \multicolumn{1}{l|}{0.745±0.061} & \textbf{0.861±0.061}      \\ \cline{2-9} 
                                          & Recall                            & \multicolumn{1}{l|}{0.731±0.113} & \multicolumn{1}{l|}{0.754±0.082} & 0.720±0.102              & \multicolumn{1}{l|}{0.455±0.157} & 0.717±0.091              & \multicolumn{1}{l|}{0.744±0.142} & \textbf{0.893±0.085}      \\ \hline
\multirow{4}{*}{\textbf{Heart failure}}   & AUC                               & \multicolumn{1}{l|}{0.779±0.027} & \multicolumn{1}{l|}{0.781±0.028} & 0.797±0.025              & \multicolumn{1}{l|}{0.871±0.031} & 0.914±0.006              & \multicolumn{1}{l|}{0.952±0.008} & \textbf{0.982±0.005}      \\ \cline{2-9} 
                                          & Precision                         & \multicolumn{1}{l|}{0.806±0.012} & \multicolumn{1}{l|}{0.786±0.023} & 0.808±0.018              & \multicolumn{1}{l|}{0.824±0.127} & 0.878±0.069              & \multicolumn{1}{l|}{0.942±0.008} & \textbf{0.979±0.009}      \\ \cline{2-9} 
                                          & F1                                & \multicolumn{1}{l|}{0.788±0.039} & \multicolumn{1}{l|}{0.793±0.034} & 0.806±0.031              & \multicolumn{1}{l|}{0.858±0.064} & 0.907±0.032              & \multicolumn{1}{l|}{0.955±0.007} & \textbf{0.982±0.008}      \\ \cline{2-9} 
                                          & Recall                            & \multicolumn{1}{l|}{0.774±0.070} & \multicolumn{1}{l|}{0.801±0.047} & 0.805±0.048              & \multicolumn{1}{l|}{0.913±0.040} & 0.941±0.013              & \multicolumn{1}{l|}{0.968±0.013} & \textbf{0.984±0.008}      \\ \hline
\end{tabular}
}
\label{mainexp}
\end{table*}

\subsection{Experiment settings}

\textbf{Dataset} We conduct our experiments on the widely used MIMIC-III~\cite{johnson2016mimic} dataset. We select $7$ kinds of diseases, including dementia~\cite{Dementia}, herpes zoster~\cite{HerpesZoster}, asthma~\cite{Asthma}, post event pain~\cite{Posteventpain}, sickle cell~\cite{Sickle}, ADHD~\cite{ADHD}, and heart failure~\cite{HF}, following previous studies \cite{de2021phe2vec}. For the patients' features, we select the related features from their medical records including ICD code, GSN code, medication and clinical notes. For the selection of negative samples, we randomly select equal patients without target disease. The dataset statistical information is presented in Table~\ref{datasetstat}.

\textbf{Baseline methods} 

To validate the effectiveness of our multimodal model, we choose three kinds of baseline methods. The first one is baseline models including logistic regression (LR)~\cite{fan2008liblinear}, multilayer perceptron (MLP) and gradient boosting classifier (GBC)~\cite{friedman2001gbdt} trained on structured EHR data. The second one is baseline models trained on clinical note data. We use the Word2Vec~\cite{mikolov2013word2vec} method to embed the clinical notes and feed the flattened embedding to baseline models including MLP and GBC for prediction. The third one is multi-modality baseline. We design a baseline multi-modal model. We concatenate the one-hot structured EHRs and clinical notes features processed by Word2Vec together, and both types of data are processed with feature selection. The concatenated features will be the inputs of GBC for prediction. 

To validate the effectiveness of the ensemble strategy, we categorize two types of baselines. The first one is our multimodal method without the ensemble strategy. The second is ensemble learning with baseline methods, for which we choose LR trained on clinical notes, GBC trained on structured EHRs and GBC trained on multi-modality data as the ensemble members.

\textbf{Evaluation protocal} For baselines and our method, we conduct $5$-fold cross validation to evaluate. We use four commonly used metrics including ROC AUC, precision, F1 score, and recall to evaluate the performance of our model and baselines. The comparisons are made across $7$ different diseases to validate the effectiveness of our method.

\textbf{Implementation details}
We implement the baselines based on Sklearn package \cite{scikit-learn}. Our model is implemented based on PyTorch package~\cite{pytorch}. We apply the open-source BlueBERT~\cite{peng2019transfer} for our BERT component. The batch size is set to $128$. Our framework is trained for $50$ epochs with Adam optimizer and $1e-3$ as the learning rate. We adopt the Snorkel ~\cite{ratner2017snorkel} as the implementation for the ensemble learning part.

\begin{table*}[h]
\caption{Ensemble learning}
\centering
\scalebox{1.1}{
\begin{tabular}{|c|c|c|cc|cc|}
\hline
\multirow{2}{*}{\textbf{Disease}} & \multirow{2}{*}{\textbf{Metrics}} & \multirow{2}{*}{\textbf{Ours w/o ensemble}} & \multicolumn{2}{c|}{\textbf{Ensemble learning of baselines}}          & \multicolumn{2}{c|}{\textbf{Ensemble learning of our model}}                 \\ \cline{4-7} 
                                  &                                   &                                             & \multicolumn{1}{c|}{Majority vote} & \multicolumn{1}{c|}{Label model} & \multicolumn{1}{c|}{Majority vote}        & \multicolumn{1}{c|}{Label model} \\ \hline
\multirow{4}{*}{\textbf{Dementia}}         & AUC                               & 0.966±0.008                                 & \multicolumn{1}{l|}{0.828±0.043}   & 0.833±0.050                      & \multicolumn{1}{l|}{\textbf{0.988±0.004}} & 0.986±0.004                      \\ \cline{2-7} 
                                  & Precision                         & 0.963±0.014                                 & \multicolumn{1}{l|}{0.772±0.037}   & 0.788±0.034                      & \multicolumn{1}{l|}{\textbf{0.987±0.004}} & 0.984±0.007                      \\ \cline{2-7} 
                                  & F1                                & 0.967±0.010                                 & \multicolumn{1}{l|}{0.852±0.016}   & 0.860±0.012                      & \multicolumn{1}{l|}{\textbf{0.989±0.002}} & 0.987±0.003                      \\ \cline{2-7} 
                                  & Recall                            & 0.971±0.013                                 & \multicolumn{1}{l|}{0.953±0.021}   & 0.947±0.024                      & \multicolumn{1}{l|}{\textbf{0.991±0.003}} & \textbf{0.991±0.005}             \\ \hline
\multirow{4}{*}{\textbf{Herpes zoster}}    & AUC                               & 0.897±0.037                                 & \multicolumn{1}{l|}{0.693±0.051}   & 0.692±0.027                      & \multicolumn{1}{l|}{\textbf{0.957±0.026}} & \textbf{0.957±0.026}             \\ \cline{2-7} 
                                  & Precision                         & 0.882±0.062                                 & \multicolumn{1}{l|}{0.650±0.112}   & 0.789±0.093                      & \multicolumn{1}{l|}{\textbf{0.963±0.025}} & \textbf{0.963±0.025}             \\ \cline{2-7} 
                                  & F1                                & 0.902±0.047                                 & \multicolumn{1}{l|}{0.727±0.120}   & 0.707±0.139                      & \multicolumn{1}{l|}{\textbf{0.965±0.024}} & \textbf{0.965±0.024}             \\ \cline{2-7} 
                                  & Recall                            & 0.924±0.044                                 & \multicolumn{1}{l|}{0.827±0.139}   & 0.717±0.265                      & \multicolumn{1}{l|}{\textbf{0.968±0.042}} & \textbf{0.968±0.042}             \\ \hline
\multirow{4}{*}{\textbf{Asthma}}           & AUC                               & 0.966±0.012                                 & \multicolumn{1}{l|}{0.779±0.036}   & 0.779±0.036                      & \multicolumn{1}{l|}{\textbf{0.984±0.005}} & 0.982±0.006                      \\ \cline{2-7} 
                                  & Precision                         & 0.966±0.012                                 & \multicolumn{1}{l|}{0.737±0.028}   & 0.737±0.028                      & \multicolumn{1}{l|}{0.981±0.006}          & \textbf{0.982±0.006}             \\ \cline{2-7} 
                                  & F1                                & 0.967±0.012                                 & \multicolumn{1}{l|}{0.812±0.021}   & 0.812±0.021                      & \multicolumn{1}{l|}{\textbf{0.984±0.006}} & 0.983±0.008                      \\ \cline{2-7} 
                                  & Recall                            & 0.969±0.016                                 & \multicolumn{1}{l|}{0.908±0.055}   & 0.908±0.055                      & \multicolumn{1}{l|}{\textbf{0.986±0.007}} & 0.984±0.010                      \\ \hline
\multirow{4}{*}{\textbf{Post event pain}}  & AUC                               & 0.922±0.030                                 & \multicolumn{1}{l|}{0.805±0.043}   & 0.787±0.055                      & \multicolumn{1}{l|}{\textbf{0.966±0.022}} & \textbf{0.966±0.022}             \\ \cline{2-7} 
                                  & Precision                         & 0.929±0.038                                 & \multicolumn{1}{l|}{0.778±0.114}   & 0.841±0.076                      & \multicolumn{1}{l|}{\textbf{0.970±0.018}} & \textbf{0.970±0.018}             \\ \cline{2-7} 
                                  & F1                                & 0.928±0.020                                 & \multicolumn{1}{l|}{0.810±0.060}   & 0.790±0.096                      & \multicolumn{1}{l|}{\textbf{0.965±0.023}} & \textbf{0.965±0.023}             \\ \cline{2-7} 
                                  & Recall                            & 0.928±0.024                                 & \multicolumn{1}{l|}{0.857±0.042}   & 0.775±0.166                      & \multicolumn{1}{l|}{\textbf{0.960±0.037}} & \textbf{0.960±0.037}             \\ \hline
\multirow{4}{*}{\textbf{Sickle cell}}      & AUC                               & 0.930±0.023                                 & \multicolumn{1}{l|}{0.768±0.055}   & 0.768±0.068                      & \multicolumn{1}{l|}{\textbf{0.968±0.023}} & \textbf{0.968±0.023}             \\ \cline{2-7} 
                                  & Precision                         & 0.921±0.053                                 & \multicolumn{1}{l|}{0.743±0.082}   & 0.778±0.062                      & \multicolumn{1}{l|}{\textbf{0.960±0.026}} & \textbf{0.960±0.026}             \\ \cline{2-7} 
                                  & F1                                & 0.930±0.020                                 & \multicolumn{1}{l|}{0.779±0.071}   & 0.766±0.107                      & \multicolumn{1}{l|}{\textbf{0.969±0.018}} & \textbf{0.969±0.018}             \\ \cline{2-7} 
                                  & Recall                            & 0.942±0.028                                 & \multicolumn{1}{l|}{0.821±0.075}   & 0.771±0.162                      & \multicolumn{1}{l|}{\textbf{0.979±0.019}} & \textbf{0.979±0.019}             \\ \hline
\multirow{4}{*}{\textbf{ADHD}}             & AUC                               & 0.857±0.074                                 & \multicolumn{1}{l|}{0.794±0.026}   & 0.717±0.052                      & \multicolumn{1}{l|}{\textbf{0.941±0.019}} & 0.938±0.018                      \\ \cline{2-7} 
                                  & Precision                         & 0.837±0.073                                 & \multicolumn{1}{l|}{0.782±0.152}   & 0.845±0.089                      & \multicolumn{1}{l|}{\textbf{0.903±0.039}} & \textbf{0.903±0.039}             \\ \cline{2-7} 
                                  & F1                                & 0.861±0.061                                 & \multicolumn{1}{l|}{0.780±0.084}   & 0.694±0.174                      & \multicolumn{1}{l|}{\textbf{0.943±0.019}} & \textbf{0.940±0.018}             \\ \cline{2-7} 
                                  & Recall                            & 0.893±0.085                                 & \multicolumn{1}{l|}{0.819±0.127}   & 0.660±0.272                      & \multicolumn{1}{l|}{\textbf{0.988±0.014}} & 0.982±0.025                      \\ \hline
\multirow{4}{*}{\textbf{Heart failure}}    & AUC                               & 0.982±0.005                                 & \multicolumn{1}{l|}{0.906±0.038}   & 0.952±0.008                      & \multicolumn{1}{l|}{\textbf{0.990±0.003}} & \textbf{0.990±0.003}             \\ \cline{2-7} 
                                  & Precision                         & 0.979±0.009                                 & \multicolumn{1}{l|}{0.867±0.013}   & 0.942±0.008                      & \multicolumn{1}{l|}{\textbf{0.990±0.004}} & \textbf{0.990±0.002}             \\ \cline{2-7} 
                                  & F1                                & 0.982±0.008                                 & \multicolumn{1}{l|}{0.921±0.010}   & 0.955±0.007                      & \multicolumn{1}{l|}{\textbf{0.990±0.005}} & \textbf{0.990±0.005}             \\ \cline{2-7} 
                                  & Recall                            & 0.984±0.008                                 & \multicolumn{1}{l|}{0.981±0.012}   & 0.968±0.013                      & \multicolumn{1}{l|}{\textbf{0.990±0.006}} & \textbf{0.989±0.008}             \\ \hline
\end{tabular}}
\label{ensembleexp}
\end{table*}

\subsection{Comparison with baseline methods}
\label{comparewithbaselines}
We make comparisons with baseline methods on $7$ different diseases (\textbf{RQ1}). As shown in Table~\ref{mainexp}, we present the results on different evaluating metrics such as ROC AUC, precision,  etc. We can observe that our model can outperform three categories of baseline models on $7$ diseases. 

Compared to single-modality data, the multi-modality baseline model can surpass the LR, MLP, GBC baselines on structured EHR by $16.63\%, 12.92\%, 15.07\%$ and the MLP, GBC baselines on clinical notes by $22.61\%, 6.71\%$ respectively on average across $7$ diseases with ROC AUC as a metric. Our multi-modality model can further outperform the structured EHR baseline models LR, MLP, GBC for $7$ diseases by an average of $33.31\%, 28.96\%, 31.44\%$ correspondingly on the ROC AUC metric. Our model have better performance compared to the clinical notes baseline models MLP and GBC with average ROC AUC improvement by $40.44\%, 21.98\%$ on $7$ diseases. These results show that multi-modality data can provide more useful information to improve performance even with the baseline multimodal method. 

Compared to the multi-modality baseline method, our method can improve the ROC AUC by $14.28\%$ on average of $7$ diseases. It implies that our proposed multimodality learning part can learn from the fusing structured EHRs and clinical notes more effectively with superior performance compared with the multi-modality baseline. 

Furthermore, we can observe that the baseline methods usually performs worse on the disease with much less patients numbers, which are herpes zoster, post event pain, sickle cell and ADHD in our experiment. But our method can still perform well on these kind of diseases. It indicates the promising application of our multi-modality method to some rare diseases~\cite{schaefer2020use} with very limited number of patients.

\subsection{Ensemble Learning Results}
\label{ensemblelearning}
Two categories of baseline methods are compared with our ensemble model (\textbf{RQ2}). As shown in Table~\ref{ensembleexp}, our model with ensemble learning can significantly outperforms our model without ensemble learning by $4.33\%$ on ROC AUC under the majority voting strategy. With the label model ensemble strategy, the multi-modality model performance can be improved by $4.22\%$ on ROC AUC. It implies the ensemble learning strategy can further boost the performance of the proposed multi-modality model. 

We also make comparisons with the baseline ensemble model. We can find that the proposed ensemble learning model with a majority voting strategy can achieve higher ROC AUC than the best baseline ensemble model with a majority voting or label model strategy by $21.65\%$. The performance of the label model ensemble strategy on our proposed framework is close to the proposed framework with majority voting. The ROC AUC has improved by $21.53\%$ averagely compared to the best baseline ensemble model of each disease. 


\subsection{Manual Chart Review}
\label{erroranalysis}

As we emphasized before, the training labels generated by PheKB algorithms are noisy and very likely to ignore some positive patients. Since our PheME has a strong learning capacity and is trained on multi-modal data, we expect the model can explore more generalized features for phenotype prediction. In order to get a better understanding of our proposed model performance, we selected a random sample of false positive cases for manual chart review. Upon review, we found three common patterns. First, patients who are pronounced dead commonly experience terminal events that our model translates into contributing factors for our prediction outcomes, such as dementia and heart failure. Second, patients might report common symptoms or risk factors for our prediction outcomes, such as shortness of breath for asthma prediction or cardiac fibrillation for heart failure risk prediction. Third, uncaptured evidence by the PheKB algorithm implementation, either for cases or controls definition. For example, as per the PheKB algorithm patients with chronic obstructive pulmonary disease (COPD) should be excluded from the control group to reduce the incidence of false positive cases. However, the control exclusion criteria in the algorithm solely rely on the structured diagnoses ICD codes. Therefore, some of the patients who do not have the ICD codes in their structured diagnoses records while having the same information in their clinical notes were captured by our model. Another example includes patients who had heart failure diagnoses documented in their clinical notes, However, the HF diagnosis was not explicitly documented in the structured EHR record. Therefore, they were wrongly labeled as controls as per the PheKB algorithm implementation while they are true cases correctly captured by our model. 


\section{Conclusion}

In this paper, we present a deep ensemble learning framework for phenotype prediction based on multi-modal data. Specifically, for sparse structured EHRs, we adopt a multi-layer perceptron to project the selected features into a dense embedding. For the clinical notes, we leverage a simple yet effective filter to select important sentences and then feed them into a language model pretrained on the medical domain to extract useful information. An ensemble learning framework combines outputs from single-modal models and multi-modal models and improve phenotype predictions. Experiments on seven diseases show that multi-modal data significantly improves phenotype prediction in all diseases, and the proposed ensemble learning technique can further improve the prediction performance, especially for diseases with limited training samples. A manual chart review shows that the proposed PheME is generalizable and can identify true positive patients that are ignored by the PheKB algorithms. In the future, we will consider improving our data preprocessing pipeline and including more data models, such as medical images and dialogue between patients and doctors. We will also consider more diseases in future work.

\bibliography{reference}
\bibliographystyle{IEEEtran}
\end{document}